# Extension of Convolutional Neural Network along Temporal and Vertical Directions for Precipitation Downscaling

Takeyoshi Nagasato[1], Kei Ishida[2], Ali Ercan[3], Tongbi Tu[4], Masato Kiyama[5], Motoki Amagasaki[6], Kazuki Yokoo[7]

[1] Graduate School of Science and Technology, Kumamoto University, 2-39-1 Kurokami, Kumamoto 860-8555, Japan

[2] International Research Organization for Advanced Science and Technology, Kumamoto University, 2-39-1 Kurokami, Kumamoto 860-8555, Japan

[3] Department of Civil and Environmental Engineering, University of California, Davis, One Shields Avenue, Davis, California 95616, USA

[4] School of Civil Engineering, Sun Yat-Sen University, Guangzhou, China

[5] Faculty of Advanced Science and Technology, Kumamoto University, 2-39-1 Kurokami, Kumamoto 860-8555, Japan

[6] Faculty of Advanced Science and Technology, Kumamoto University, 2-39-1 Kurokami, Kumamoto 860-8555, Japan

[7] Graduated School of Science and Technology, Kumamoto University, 2-39-1 Kurokami, Kumamoto 860-8555, Japan

**Corresponding Author**: Kei Ishida, International Research Organization for Advanced Science and Technology, Kumamoto University, 2-39-1 Kurokami, Kumamoto 860-8555, Japan

Email: keiishida@kumamoto-u.ac.jp

**Abstract**

Deep learning has been utilized for the statistical downscaling of climate data. Specifically, a two-dimensional (2D) convolutional neural network (CNN) has been successfully applied to precipitation estimation. This study implements a three-dimensional (3D) CNN to estimate watershed-scale daily precipitation from 3D atmospheric data and compares the results with those for a 2D CNN. The 2D CNN is extended along the time direction (3D-CNN-Time) and the vertical direction (3D-CNN-Vert). The precipitation estimates of these extended CNNs are compared with those of the 2D CNN in terms of the root-mean-square error (RMSE), Nash-Sutcliffe efficiency (NSE), and 99$^{th}$ percentile RMSE. It is found that both 3D-CNN-Time and 3D-CNN-Vert improve the model accuracy for precipitation estimation compared to the 2D CNN. 3D-CNN-Vert provided the best estimates during the training and test periods in terms of RMSE and NSE.

**Keywords:** Precipitation downscaling, Watershed-scale precipitation, Convolutional neural network, Atmospheric data, ERA-Interim

# 1. INTRODUCTION

Meteorological disasters have become more intense and more frequent in many parts of the world, likely due to global warming. Therefore, the effects of global warming on meteorological variables, including its future effects, must be taken into account. Most studies on these effects use future climate projections simulated using general circulation models based on emission scenarios. However, the spatial resolution of future climate projections is often more than 100 km, which is insufficient for regional- or watershed-scale analysis. Therefore, downscaling is frequently employed to obtain meteorological variables at a finer resolution.

There are two major downscaling techniques, namely dynamical downscaling and statistical downscaling (Bhuvandas *et al.*, 2014). Dynamical downscaling utilizes a physically based three-dimensional (3D) model, called a regional climate model, which uses variables from future climate projections as its initial and boundary conditions, and then generates target variables at a finer resolution from numerical simulations. Statistical downscaling obtains variables at a finer resolution based on consideration of the statistical relations between observation data and the historical runs of climate models. There are various statistical downscaling methods. They can be categorized into three types, namely regression methods, weather-pattern-based methods, and weather generators (Wilby and Wigley, 1997). Details of statistical downscaling methods can be found in the review papers of Wilby and Wigley (1997), Schoof (2013), and Maraun *et al.* (2010).

Deep learning has been utilized for climate data downscaling. For example, Vandel *et al.* (2017) utilized a convolutional neural network (CNN), a deep learning method that can capture two-dimensional (2D) features from images or 2D data fields, to generate fine-resolution precipitation fields from coarse-resolution precipitation data. Pan *et al.* (2019) utilized a CNN to estimate the daily precipitation in a single grid cell from atmospheric data. As atmospheric data, they utilized the geopotential height field on several isobaric surfaces and the precipitable water field obtained from an atmospheric reanalysis dataset. Similarly, Nagasato *et al.* (2020) utilized a CNN to estimate the daily basin-average precipitation at a watershed from atmospheric data, and investigated the effects of input variables on the accuracy of precipitation estimation.

Misra *et al.* (2018) employed another deep learning method, namely a long short-term memory (LSTM) network, to estimate precipitation at gauging stations from meteorological data obtained from a reanalysis dataset. LSTM is a type of recurrent neural network (RNN) that is suitable for sequential data. Tran Anh *et al.* (2019) utilized atmospheric data from general circulation models. Miao *et al.* (2019) utilized a combined model of a CNN and LSTM to consider the 2D and sequential features of atmospheric data obtained from a reanalysis dataset. Then, they compared the combined model with a CNN for precipitation estimation at gauging stations. Sadeghi *et al.* (2019) utilized a CNN for estimate precipitation from infrared and water vapor satellite data.

These deep-learning-based precipitation downscaling studies obtained successful results. However, deep learning methods have rapidly evolved. There are many newer deep learning methods that may be more suitable for downscaling and may improve the accuracy of precipitation estimation. RNNs, including LSTM, are computationally expensive due to their recurrent structures (Kratzert *et al.*, 2018). The combination of a CNN and an RNN (LSTM) can be used to capture the transitions of atmospheric data in addition to 2D features. However, the use of 2D information by an RNN requires a lot of computational resources (Shi *et al.*, 2015). A CNN can be extended to more than two dimensions. 3D CNNs have been utilized to deal with video images (Ji *et al.*, 2013) and to read the time series of precipitation fields and forecast precipitation fields several time steps ahead (Shi *et al.*, 2017).

Within the above framework, this study implements a 3D CNN to estimate watershed-scale daily precipitation from 3D atmospheric data and compares the results with those obtained using a 2D CNN. Previous studies have utilized a 3D CNN to capture the sequential features of data in addition to their 2D features. This study extends the direction of the CNN along the time direction in addition to the horizontal direction. This CNN is referred to as 3D-CNN-Time. The vertical expansion of atmospheric conditions may also be important for estimating precipitation. Therefore, this study also extends the direction of the CNN along the vertical direction. This CNN is referred to as 3D-CNN-Vert. Similar to previous studies (Misra *et al.*, 2018; Miao *et al.*, 2019; Pan *et al.*, 2019; Nagasato *et al.*, 2020), this study utilizes a reanalysis dataset to obtain atmospheric data. The Shira River watershed, in southern Japan, was selected as the study watershed. The daily basin-average precipitation over the study watershed was utilized as the target data. After the 2D CNN, 3D-CNN-Time, and 3D-CNN-Vert were tuned, the model accuracy of 3D-CNN-Time and 3D-CNN-Vert were compared with that of the 2D CNN in terms of several statistical evaluation metrics.

## 2. METHODOLOGY

This study compares 2D and 3D CNNs for precipitation downscaling. A CNN is composed of convolutional, pooling, and fully connected layers. The convolutional and pooling layers of 2D and 3D CNNs receive 2D and 3D data as inputs, respectively. Each input is counted as a channel. Each layer extracts features from the inputs via filters. The extracted feature data and the filters are called feature maps and kernels, respectively. Each feature map is counted as a channel. The dimensions of the inputs, kernels, and feature maps must match for a given CNN.

Convolutional and pooling layers perform different operations. The kernels of a convolutional layer work as learnable parameters (weights). Each grid of a kernel has a value. These values are optimized through a training process. The number of kernels needs to be set for each convolutional layer. The convolution operation scans each channel of the input from one corner to the opposite corner using a kernel, and then generates a channel of the feature map. As shown in Figure 1, this scanning process for convolution operation is conducted in two and three dimensions for 2D and 3D CNNs, respectively. The convolution operation is repeated as many times as the number of kernels. Consequently, the number of channels of the feature map is defined by the number of kernels in the convolution operation. The shift amount, called the stride, in the scanning process can be adjusted. The convolution operation may lose information at the edges of the inputs. To avoid this, the inputs are frequently enveloped by zero, a process known as padding.

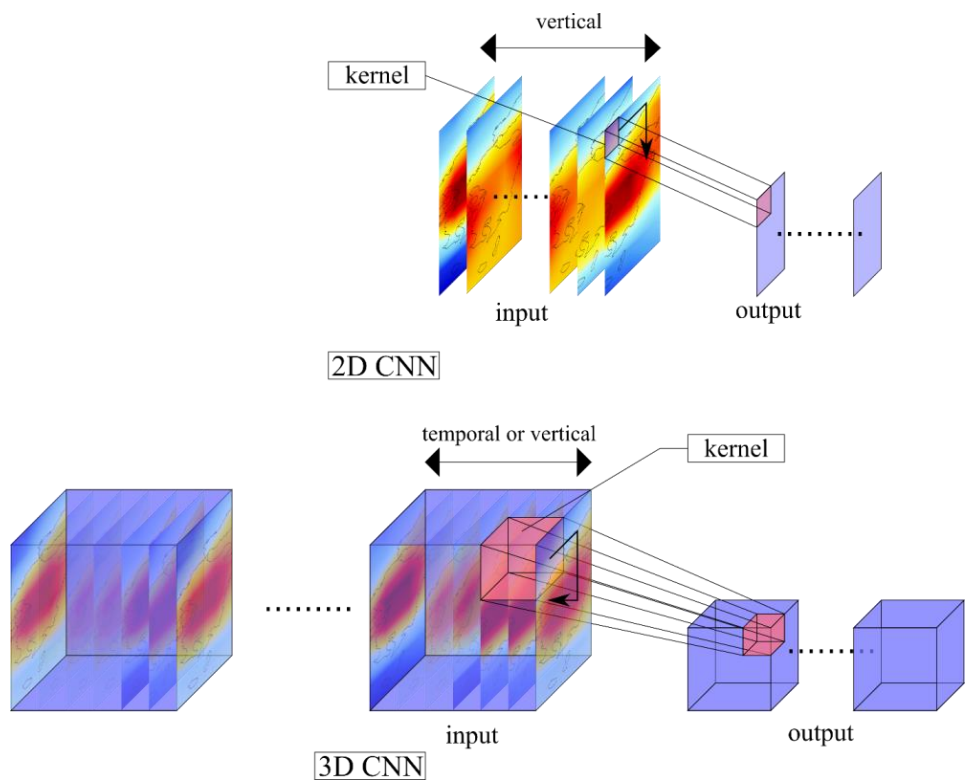

**Figure 1.** Scanning process of convolution operation.

The convolution operation for the 2D CNN is written as follows:

$$v_{ij}^p = \sum_{m=1}^{M} \sum_{s=0}^{S-1} \sum_{t=0}^{T-1} k_{st}^p u_{(i+s)(j+t)}^m + b^p$$

where $v$ is the feature map, $k$ is the kernel, $u$ is the input, $b$ is the bias, and $p$ is the number of kernels, which is equal to the channel number of the feature map. $S$ and $T$ are the 2D sizes of the kernels. $M$ is the channel number of the input. $i$ and $j$ represent the position in the feature map and are linked to the position of the input. The increment and range of $i$ and $j$ depend on the stride and padding width, respectively. The bias $b$ is a learnable parameter. The 3D convolution operation is similar to the 2D one. It just has an additional dimension. The 3D convolution operation is defined as

$$v_{ijn}^p = \sum_{m=1}^{M} \sum_{s=1}^{S-1} \sum_{t=0}^{T-1} \sum_{w=0}^{W-1} k_{stw}^p u_{(i+s)(j+t)(n+w)}^m + b^p$$

where $S$, $T$, and $W$ are the 3D sizes of the kernels, and $i$, $j$, and $n$ represent the 3D position in the feature map.

The pooling operation extracts features from the input channels via a simple statistic. Similar to the convolution operation, the pooling operation scans the input channels using a kernel. However, the kernel of the pooling operation does not work as a learnable parameter. As the pooling operation shifts the kernel from one corner to the opposite corner, it calculates a statistical value over the kernel's covered area at each position. Because the pooling operation generates a channel of the feature map from a channel of the input, the input and the feature map have the same number of channels. The only difference in the pooling operation between 2D and 3D CNNs is the dimension of the input, kernel, and feature map. There are several types of pooling operation based on the statistics used. This study utilizes max pooling, which extracts the maximum value.

A fully connected layer receives an input vector and then generates an output vector generally via an affine transformation. The size of the input vector is given, whereas that of the output vector needs to be set. The matrix and bias of the affine transformation are learnable weight parameters of a CNN, in addition to the kernels of the convolutional layers. The input vector of the first fully connected layer is the output feature maps of the last pooling layer (Figure 2). The input vector of the first fully connected layer is one-dimensional whereas the feature map consists of multiple channels of multi-dimensional data. Therefore, the feature maps are flattened to a single vector before being given to the first fully connected layer.

The mini-batch gradient descent method is generally utilized for the training process in deep learning. For this method, the training dataset is divided into subsets, called mini-batches or batches. The training process is conducted with each mini-batch. During the training process with the mini-batch gradient descent method, the distribution of the values of the input to the layers except the first layer changes because the learnable parameters for each convolutional layer change. This change in the distribution of inputs to layers in the network is called an internal covariate shift, which disturbs training. Batch normalization was developed to solve this problem (Ioffe and Szegedy, 2015) by normalizing the distribution of the values of the input. With batch normalization, each input in mini-batch $B = \{x_1 \dots x_m\}$ is normalized by the mean and variance of the inputs:

$$y_i = \gamma \frac{x_i - \mu_B}{\sqrt{\sigma_B^2 + \epsilon}} + \beta$$

where $\mu_B$ is the mini-batch mean and $\sigma_B^2$ is the mini-batch variance. $\epsilon$ is a constant used for numerical stability and $\gamma$ and $\beta$ are learnable parameters.

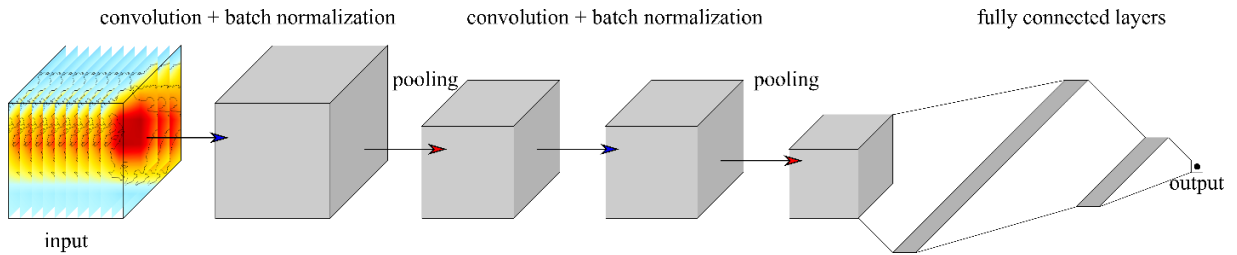

**Figure 2.** Structure of a CNN.

2D and 3D CNNs receive 2D and 3D data, respectively, as input. This study uses the time series of 3D atmospheric data as input and estimates the basin-average precipitation at the study watershed. The 3D atmospheric time series data need to be divided into channels to be given to a CNN. The 3D atmospheric time series data for each variable are separated into 2D data on each isobaric surface at each time for the 2D CNN. This study utilizes two 3D CNNs, namely 3D-CNN-Time and 3D-CNN-Vert. For 3D-CNN-Time, the data are separated into 3D data that are spread on an isobaric surface and along time. For 3D-CNN-Vert, the data are divided into time steps, resulting in 3D spatial data at each time. After the given data pass through all the layers and operations, the CNNs generate a single value of the basin-average precipitation at each time.

## 3. APPLICATION

### 3.1. Study Area

The Shira River Basin (SRB), Kyushu, Japan, was selected as the study area (Figure 3). The length of Shira River is 74 km and the area of the SRB is 480 km$^2$. The upstream of Shira River is located in Aso caldera, which occupies 80% of the SRB. Additionally, Shira River has a raised bed downstream.

In general, heavy precipitation due to a Baiu front occurs in June and July over the Kyushu region and typhoons hit the region from July to November. Thus, heavy rain falls from June to November in the SRB. In July 2017, there was record-breaking heavy rain in northern Kyushu, with 183 houses completely or partially destroyed. The precipitation was recorded as 124 mm per hour. There is not much snowfall in the Kyushu region. The SRB is prone to severe flood damage because of the geographical and meteorological conditions.

### 3.2. Dataset

This study uses atmospheric reanalysis data as the input and the observed precipitation as the target data. The observed precipitation data were obtained from the Asian Precipitation - Highly Resolved Observational Data Integration Towards Evaluation Data of Water Resources (APHRODITE). APHRODITE (Kamiguchi *et al.*, 2010) is a gridded daily precipitation dataset based on gauging observations. It covers all of Japan with a spatial resolution of 0.05° × 0.05° (about 5 km) from the years 1900 to 2015. The basin-average values at the study watershed were calculated from the gridded precipitation data.

The ERA-Interim atmospheric reanalysis data from the European Centre for Medium-Range Weather Forecasts (ECMWF) was selected as the reanalysis source of data (Dee *et al.*, 2011). ERA-Interim spatially covers the entire world and temporally ranges from 1 January, 1979, to August 2019. It provides various 3D atmospheric variables at 37 pressure levels from 1 to 1000 hPa. The horizontal resolution is 0.75° × 0.75° (about 80 km) and the temporal resolution is 6 hours (0:00, 6:00, 12:00, and 18:00 UTC). This study selected the following 3D variables on isobaric surfaces: specific humidity, meridional wind speed, zonal wind speed, vertical wind speed, and air temperature. These variables are similar to those in a study by Nagasato *et al.* (2020), who used a 2D CNN for precipitation downscaling. The horizontal range of the atmospheric variables used in this study is shown in Figure 4.

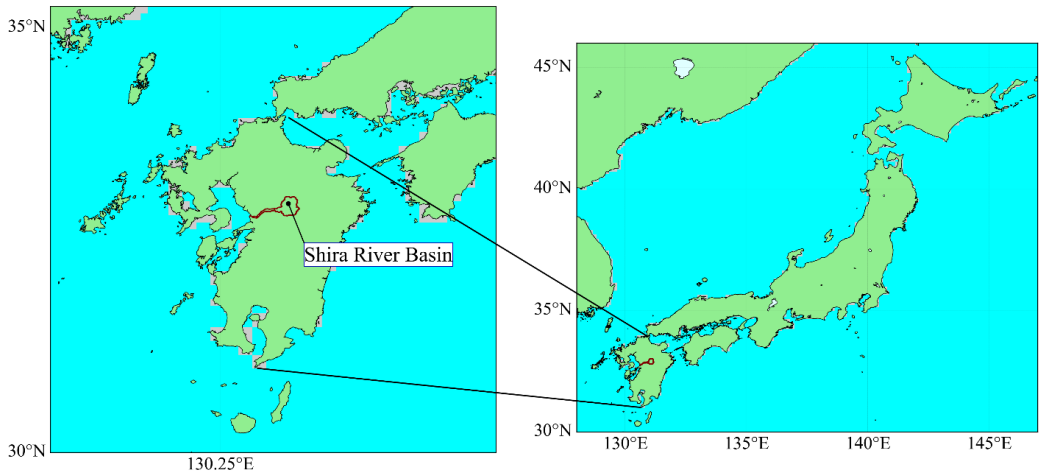

**Figure 3.** Shira River Basin, Kyushu, Japan.

### 3.3. Model Implementation

This study compares 3D-CNN-Time and 3D-CNN-Vert with the 2D CNN for precipitation downscaling. The CNNs were forced by the atmospheric time series data on isobaric surfaces obtained from ERA-Interim. The basin-average precipitation of the study watershed (480 km$^2$) was selected as the target data. Because a single time series was targeted, this process is not technically precipitation downscaling. The model described below can be easily extended to estimate precipitation at multiple grids over a region or watershed of interest.

The structure of the 2D CNN used in this study consists of two sets of convolutional layers, a max-pooling layer, and two fully connected layers, as shown in Figure 2. After each convolutional layer, batch normalization was applied (Ioffe and Szegedy, 2015). The 3D CNNs had the same structure as that of the 2D CNN, but the convolution and max-pooling operations were 3D. For the implementation, this study employed PyTorch, which is a deep learning framework written in Python.

The temporal resolution of ERA-Interim data is 6 hours. There is a 9-hour time gap between UTC and Japanese Standard Time. In Japan time, the four time steps are 3:00, 9:00, 15:00, and 21:00. This study considered three time step combinations: two time steps, four time steps, and six time steps. The two time steps were 3:00 and 15:00 within a day (corresponding to the daily precipitation), the four time steps were 3:00, 9:00, 15:00, and 21:00 within a day, and the six time steps were 21:00 on the previous day and 3:00 on the next day in addition to the four time steps within a day.

Three sets of isobaric surfaces were considered to obtain 3D atmospheric data. The first set of data was obtained from five isobaric surfaces, namely 500, 700, 850, 925, and 1000 hPa. The second set contained 300 hPa in addition to the first set. The third set consisted of 200, 300, 400, 500, 600, 700, 750, 800, 850, 900, 950, and 1000 hPa. The combinations of time steps and isobaric surfaces used for the comparisons are tabulated in Tables 1 and 2. A comparison between the 2D CNN and 3D-CNN-Vert was conducted with the three sets of isobaric surfaces. That between the 2D CNN and 3D-CNN-Time was conducted with the three combinations of time steps. The number of input channels (NC) given to the first layer of a CNN is shown in Tables 1 and 2. Even when the same number of isobaric surfaces and time steps are used, the number of input channels is different between 2D and 3D CNNs, as explained in the methodology section.

The kernel size and the stride were set to three and one, respectively, for all convolutional and pooling layers of all CNNs. The mini-batch gradient descent method was utilized, which updates the learnable parameters of the CNN using the subsets of the training datasets. The subsets are called batches. The gradient descent method requires the calculation of errors, called losses, for the update. This study uses the mean square error (MSE) as the loss function, expressed as:

$$\text{MSE} = \frac{1}{N}\sum_{i=1}^{N}(E_i - y_i)^2$$

where $y$ is the true value, $E$ is the estimated value, and $N$ is the number of data. The learning rate for the mini-batch gradient descent method was adjusted using the Adam optimization algorithm (Kingma and Ba, 2015). The batches were generated using the batch shuffling method, which randomly extracts data from the training dataset without any duplication. The batch size was set to 512. Updating with all batches was counted as an epoch.

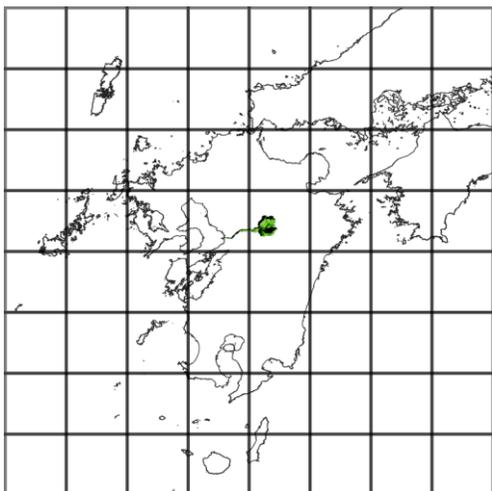

**Figure 4.** Horizontal range of the atmospheric variables used in this study.

**Table 1.** Combinations of time steps and isobaric surfaces for comparisons of 2D CNN and 3D-CNN-Vert (NC: number of input channels).

| Case | | Algorithm | # of time steps | Grid size | NC |
|---|---|---|---|---|---|
| V1 | V1-2D | 2D CNN | 6 | 5 (500/700/850/925/1000 hPa) | 120 |
|  | V1-3D | 3D-CNN-Vert | 6 | 5 (500/700/850/925/1000 hPa) | 24 |
| V2 | V2-2D | 2D CNN | 6 | 6 (300/500/700/850/925/1000 hPa) | 144 |
|  | V2-3D | 3D-CNN-Vert | 6 | 6 (300/500/700/850/925/1000 hPa) | 24 |
| V3 | V3-2D | 2D CNN | 6 | 12 (200/300/400/500/600/700/750/800/850/900/950/1000 hPa) | 288 |
|  | V3-3D | 3D-CNN-Vert | 6 | 12 (200/300/400/500/600/700/750/800/850/900/950/1000 hPa) | 24 |

**Table 2.** Combinations of time steps and isobaric surfaces for comparisons of 2D CNN and 3D-CNN-Time (NC: number of input channels).

| Case | | Algorithm | # of time steps | Grid size | NC |
|---|---|---|---|---|---|
| T1 | T1-2D | 2D | 2 | 5 (500/700/850/925/1000 hPa) | 40 |
|  | T1-3D | 3D-Time | 2 | 5 (500/700/850/925/1000 hPa) | 20 |
| T2 | T2-2D | 2D | 4 | 5 (500/700/850/925/1000 hPa) | 80 |
|  | T2-3D | 3D-Time | 4 | 5 (500/700/850/925/1000 hPa) | 20 |
| T3 | T3-2D | 2D | 6 | 5 (500/700/850/925/1000 hPa) | 120 |
|  | T3-3D | 3D-Time | 6 | 5 (500/700/850/925/1000 hPa) | 20 |

At each epoch, the CNN with the updated learnable parameters was run with the validation dataset to check for overfitting. In this study, overfitting is defined as the loss obtained with the validation dataset continuously increasing during more than 40 epochs. Once overfitting was detected, updating with the training dataset was terminated. Then, the learnable parameters that showed the best results in terms of losses for the validation dataset were selected as the trained results.

The trained results are affected by the initial states of the learnable parameters, which are generally given by random values. Therefore, the above training processes with the training and validation datasets were conducted 200 times. Finally, the best trained parameters were selected among 200 trained results in terms of losses during the validation dataset as the final trained parameters. The CNN with the final trained parameters was tested with the test dataset. Because of the availability of the precipitation data and the reanalysis data, this study targets a 36-year period from 1980 to 2015. The dataset was divided into three periods to create the training (1980-2005), validation (2006-2010), and test datasets (2011-2015).

This study compares 3D-CNN-Time and 3D-CNN-Vert with the 2D CNN. For the comparison of model accuracy, this study selected the Nash-Sutcliffe efficiency (NSE) (Nash and Sutcliffe, 1970) and the root-mean-square error (RMSE). NSE and RMSE are defined as follows:

$$\text{NSE} = 1 - \frac{\sum_{t=1}^{T}(P_{obs}^t - P_{sim}^t)^2}{\sum_{t=1}^{T}(P_{obs}^t - \overline{P_{obs}})^2}$$

$$\text{RMSE} = \sqrt{\frac{1}{N}\sum_{i=1}^{N}(E_i - y_i)^2}$$

where $T$ denotes the calculation time, $P_{obs}^t$ and $P_{sim}^t$ denote the observed precipitation and estimated flow rate at time $t$, respectively, and $\overline{P_{obs}}$ denotes the average value of the observed flow rate. $y$ is the true value, $E$ is the estimated value, and $N$ is the number of data. In addition to these evaluation indices, the observed precipitation data at the 99th percentiles and the RMSE at the estimated value at that time were calculated for evaluating the estimation accuracy at the peak. This is defined as 99th percentile RMSE (RMSE99) in this study. Here, the 99th percentile value of the observed precipitation is 99.9, 105, and 102 mm in the training, verification, and test periods, respectively.

## 4. RESULTS

### 4.1. 3D-CNN-Time

The best model in terms of the minimum loss for the validation period was extracted for each configuration. Then, the three evaluation metrics, RMSE, NSE, and RMSE99, for the three periods were calculated. The results are tabulated in Table 3. When the number of used time steps was set to two (T1), all three evaluation metrics for 3D-CNN-Time were better than those for the 2D CNN for all three periods. For instance, 3D-CNN-Time improved RMSE, NSE, and RMSE99 by 0.4 mm, 0.021, and 0.5 mm, respectively, for the test period. When six time steps were used as inputs (T3), 3D-CNN-Time mostly showed better evaluation metrics compared to the 2D CNN. 3D-CNN-Time improved RMSE and NSE by -1.7 mm and 0.078 for the training period, 0.3 mm and 0.014 for the validation period, 0.3 mm and 0.020 for the test period, respectively. RMSE99 was largely improved by 3D-CNN-Time for the training and validation periods. The improvements from the 2D CNN are 15.6 and 11.9 mm for the two periods, respectively. Although RMSE99 for the test period for 3D-CNN-Time is worse than that for the 2D CNN, the difference is only 0.3 mm. In contrast, 3D-CNN-Time shows worse RMSE99 values than those for the 2D CNN for all three periods in T2. RMSE and NSE for 3D-CNN-Time are also worse than those for the 2D CNN for the training period although they are better for the validation and test periods. RMSE and NSE are 0.45 mm and 0.02 worse for 3D-CNN-Time than those for the 2D CNN for the test period, but 0.2 mm and 0.012 better, respectively, for the validation period, and 0.6 mm and 0.032 better, respectively, for the test period.

For each CNN, the use of more time steps as inputs mostly improved the model accuracy. The evaluation metrics for the test period gradually improved as more time steps were used for both CNNs. For instance, NSE for the 2D CNN was 0.595, 0.618, and 0.649 for two, four, and six time steps, respectively (Table 3). For 3D-CNN-Time, it was 0.616, 0.650, and 0.669, respectively. Clear relations between the number the used time steps could not be found in the evaluation metrics among the used time steps for both CNNs. Nevertheless, the best accuracy in terms of the three metrics was found with the use of six time steps for 3D-CNN-Time. Consequently, 3D-CNN-Time with six time steps showed the best evaluation metrics except RMSE99 for the test period among the three cases for the two methods.

Figures 5 show the time series of the simulated precipitation and the corresponding daily observed precipitation in cases T1-T3 for the 2D CNN and 3D-CNN-Time. Based on the evaluation metrics discussed above, all simulated precipitation values generally show good agreement with the observed ones. However, the time series plots do not accurately replicate the peaks of daily precipitation, especially for the test period. Many of the large peaks were underestimated by both CNNs in all cases. As shown in Table 3, RMSE99 is in the ranges of 45.9-62.3, 45.9-62.5, and 83.2-90.6 for the training, validation, and test periods, respectively. RMSE99 for the test period is much worse than that for the other periods. Figure 6 shows a one-to-one plot of the simulated and observed precipitations during the test period. The differences in the peaks between the two CNNs and among the cases are difficult to visually distinguish. Observed precipitations of more than 100 mm were mostly underestimated by the simulations. Especially, the largest peak was significantly underestimated by all simulations.

**Table 3.** Estimated accuracy in various periods when loss was the lowest in the validation period for 2D CNN and 3D-CNN-Time.

|    | Case  | Training |       |        | Validation |       |        | Test |       |        |
|----|-------|------|-------|--------|------|-------|--------|------|-------|--------|
|    |       | RMSE | NSE   | RMSE99 | RMSE | NSE   | RMSE99 | RMSE | NSE   | RMSE99 |
| T1 | T1-2D | 9.02 | 0.797 | 48.8   | 10.8 | 0.720 | 62.5   | 13.2 | 0.595 | 90.6   |
|    | T1-3D | 8.98 | 0.799 | 48.5   | 10.6 | 0.729 | 58.2   | 12.8 | 0.616 | 90.1   |
| T2 | T2-2D | 8.90 | 0.802 | 45.9   | 10.4 | 0.739 | 53.3   | 12.8 | 0.618 | 84.3   |
|    | T2-3D | 9.35 | 0.782 | 56.8   | 10.2 | 0.751 | 56.3   | 12.2 | 0.650 | 89.0   |
| T3 | T3-2D | 10.2 | 0.742 | 62.3   | 10.2 | 0.750 | 57.8   | 12.2 | 0.649 | 83.2   |
|    | T3-3D | 8.50 | 0.820 | 46.7   | 9.90 | 0.764 | 45.9   | 11.9 | 0.669 | 83.5   |

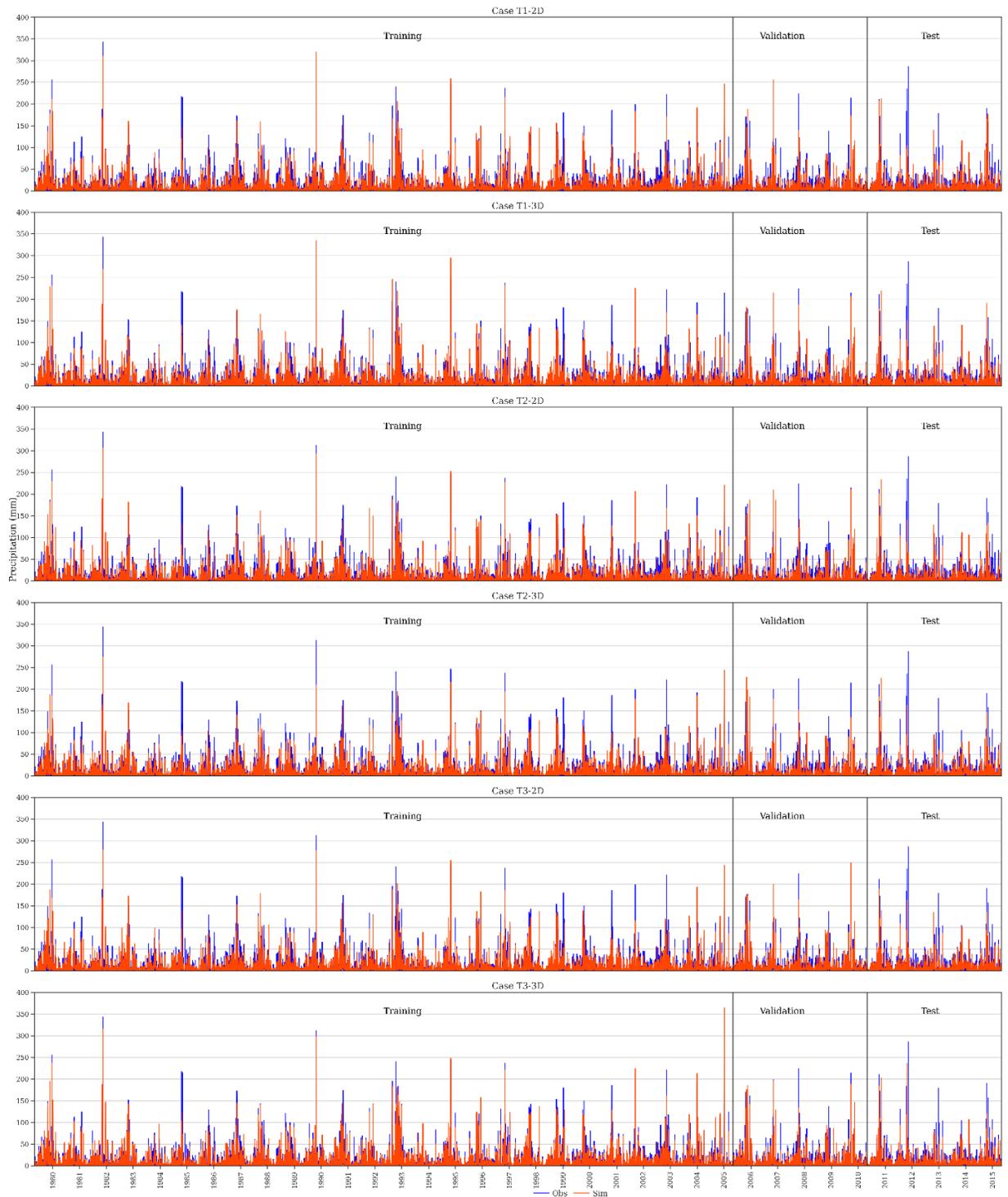

**Figure 5.** Time series plots of estimated (red) and observed (blue) values during the test period when loss was the lowest during the validation period for 2D CNN and 3D-CNN-Time.

### 4.2. 3D-CNN-Vert

Compared to 3D-CNN-Time, 3D-CNN-Vert more consistently improved model accuracy. As tabulated in Table 4, 3D-CNN-Vert shows better evaluation metrics than those of the 2D CNN except RMSE and NSE in caseT3 for the validation period. When five isobaric levels were used (V1), 3D-CNN-Vert improved RMSE, NSE, and RMSE99 by 0.8 mm, 0.037, and 7.9 mm, for the training period, 0.38 mm, 0.018, and 4.6 mm, for the validation period, and 0.5 mm, 0.032, and 5.8 mm for the test period, respectively. When six isobaric levels were used (V2), the improvements in RMSE, NSE, and RMSE99 by 3D-CNN-Vert were 0.56 mm, 0.026, and 3.6 mm for the training period, 0.2 mm, 0.009, and 2.8 mm for the validation period, and 0.1 mm, 0.009, and 6.3mm for the test period, respectively. The three evaluation metrics, especially RMSE99, for 3D-CNN-Vert are much better than those for the 2D CNN for the training period. The improvements in RMSE, NSE, and RMSE99 are 1.2 mm, 0.047, and 16.0 mm, respectively. For the test period, 3D-CNN-Vert improved the metrics by 0.3 mm, 0.020, and 5.7 mm, respectively. Although RMSE and NSE for 3D-CNN-Vert are worse than those for the 2D CNN for the validation period, the differences are only 0.1 mm and 0.002, respectively. Thus, 3D-CNN-Vert more consistently improved model accuracy than did 3D-CNN-Time.

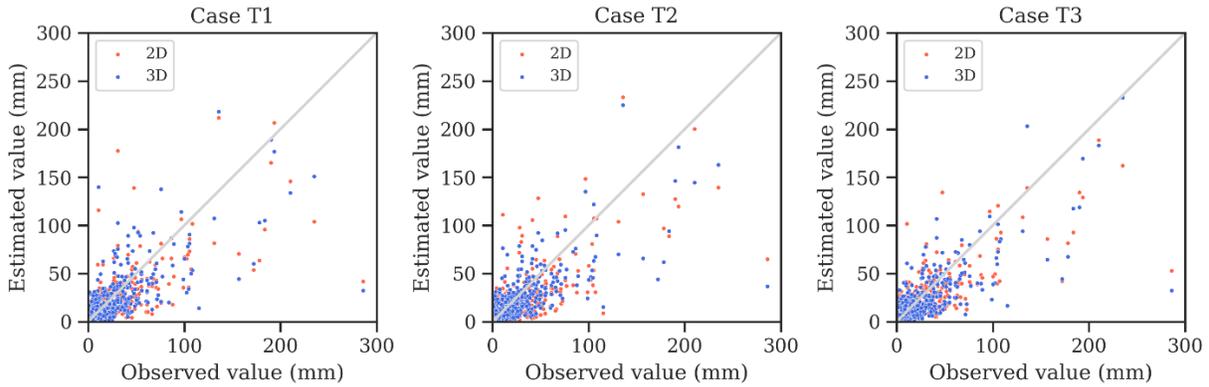

**Figure 6.** Scatter plots of estimated and observed values during the test period when loss was the lowest during the validation period for 2D CNN and 3D-CNN-Time.

**Table 4.** Estimated accuracy in various periods when loss was the lowest in the validation period for 2D CNN and 3D-CNN-Vert.

|     | Case   | Training |       |        | Validation |       |        | Test |       |        |
|-----|--------|------|-------|--------|------|-------|--------|------|-------|--------|
|     |        | RMSE | NSE   | RMSE99 | RMSE | NSE   | RMSE99 | RMSE | NSE   | RMSE99 |
| V1  | V1-2D  | 10.2 | 0.742 | 62.3   | 10.2 | 0.750 | 57.8   | 12.2 | 0.649 | 83.2   |
|     | V1-3D  | 9.40 | 0.779 | 54.4   | 9.82 | 0.768 | 53.2   | 11.7 | 0.681 | 77.4   |
| V2  | V2-2D  | 9.68 | 0.766 | 58.1   | 10.3 | 0.744 | 60.4   | 12.1 | 0.656 | 85.4   |
|     | V2-3D  | 9.12 | 0.792 | 54.5   | 10.1 | 0.753 | 57.6   | 12.0 | 0.665 | 79.1   |
| V3  | V3-2D  | 8.40 | 0.824 | 44.2   | 10.1 | 0.752 | 58.0   | 11.9 | 0.667 | 83.8   |
|     | V3-3D  | 7.20 | 0.871 | 28.2   | 10.2 | 0.750 | 57.2   | 11.6 | 0.687 | 78.1   |

In contrast, the improvements obtained by using more isobaric levels were not consistent. The 2D CNN improved RMSE and NSE when the number of isobaric levels was increased for the training and test periods. However, the best result for the 2D CNN was in case V1 for the validation period. 3D-CNN-Vert obtained the best model accuracy in case V3 for the training and test periods. However, case V3 is the worst for the validation period in terms of RMSE and NSE. Case V1 is the best for the validation period but worse for the training period in terms of RMSE and NSE. 3D-CNN-Vert consistently improved model accuracy compared to the 2D CNN, but it did not consistently improve it when the number of isobaric levels was increased.

Figures 7 show the time series of the simulated precipitation and the corresponding observed daily precipitation in cases V1-V3 for the 2D CNN and 3D-CNN-Vert. Similar to the previous section (cases T1-T3 and Figure 6), the simulated precipitation is generally in good agreement with the corresponding observations in all three cases for both CNNs. However, the large peaks were clearly underestimated except during the training period in case T3 for 3D-CNN-Vert. The peaks of the simulated precipitation also fit the observations well for the training period in case V3 for 3D-CNN-Vert, as indicated by the low RMSE99. Figure 8 shows a one-to-one plot of the simulated and observed precipitations for the test period. RMSE99 for 3D-CNN-Vert is better than that for the 2D CNN in all three cases for the test period. However, the improvements in the peaks are not consistent. Similar to cases T1-T3 (Figure 6), it is difficult to detect the improvements by a visual comparison.

## 5. DISCUSSION

This study implemented 3D-CNN-Time and 3D-CNN-Vert to estimate watershed-scale precipitation at a small watershed from 3D atmospheric reanalysis time series data. The results of the 3D CNNs were compared with those of a 2D CNN in terms of RMSE, NSE, and RMSE99. The results show the potential of both 3D-CNN-Time and 3D-CNN-Vert in improving model accuracy for precipitation estimation compared to the 2D CNN. 3D-CNN-Vert more consistently improved model accuracy than did 3D-CNN-Time (Tables 3 and 4). The evaluation metrics were improved by 3D-CNN-Vert in the three cases with a few exceptions. Especially for the test period, all evaluation metrics for 3D-CNN-Vert were better than those for the 2D CNN. In contrast, the improvements in the evaluation metrics by 3D-CNN-Time were relatively limited. 3D-CNN-Time may even worsen RMSE99.

The dimensions of 3D-CNN-Time are two horizontal dimensions and time and those of 3D-CNN-Vert are the three spatial dimensions. The vertical expansions and variations of atmospheric variables affects precipitation processes. The improvements by 3D-CNN-Vert indicate that this CNN has the potential to capture such vertical expansions and variations. 3D CNNs are frequently used for 2D image sequences (e.g., Ji *et al.*, 2013) and 2D time series data (e.g., Shi *et al.*, 2017), whose dimensions are two spatial dimensions and time. These studies showed that 3D CNNs have the potential to capture 3D features from data with two spatial dimensions and one temporal dimension. 3D-CNN-Time improved the accuracy of precipitation estimation, but the improvements were more limited than those by 3D-CNN-Vert. This study employed ERA-Interim to extract atmospheric variables. Its temporal resolution is 6 hours. This temporal resolution may be insufficient to capture the temporal expansion of atmospheric variables. A newer reanalysis dataset, ERA5 (Hersbach *et al.*, 2020), has a 1-hour temporal resolution and a 0.25° horizontal resolution. The use of ERA5 to extract atmospheric time series data for inputs may help 3D-CNN-Time improve model accuracy compared to a 2D CNN although such finer resolution data requires greatly more computational resources.

The results of this study also indicate that a CNN should be extended along the vertical direction as well as to the temporal direction to improve the accuracy of precipitation estimation using atmospheric data. A CNN can be technically extended to four dimensions, referred to as a 4D CNN. A 4D CNN has been successfully applied (e.g., Choy *et al.*, 2019). The use of 4D CNNs may also improve the accuracy of precipitation estimation. Miao *et al.* (2019) utilized the combined model of a 2D CNN and LSTM for precipitation estimation to consider the temporal expansion of the atmospheric data in addition to the horizontal expansion. They showed the high capability of the combined model for precipitation estimation. If 3D-CNN-Vert is combined with LSTM, the combined

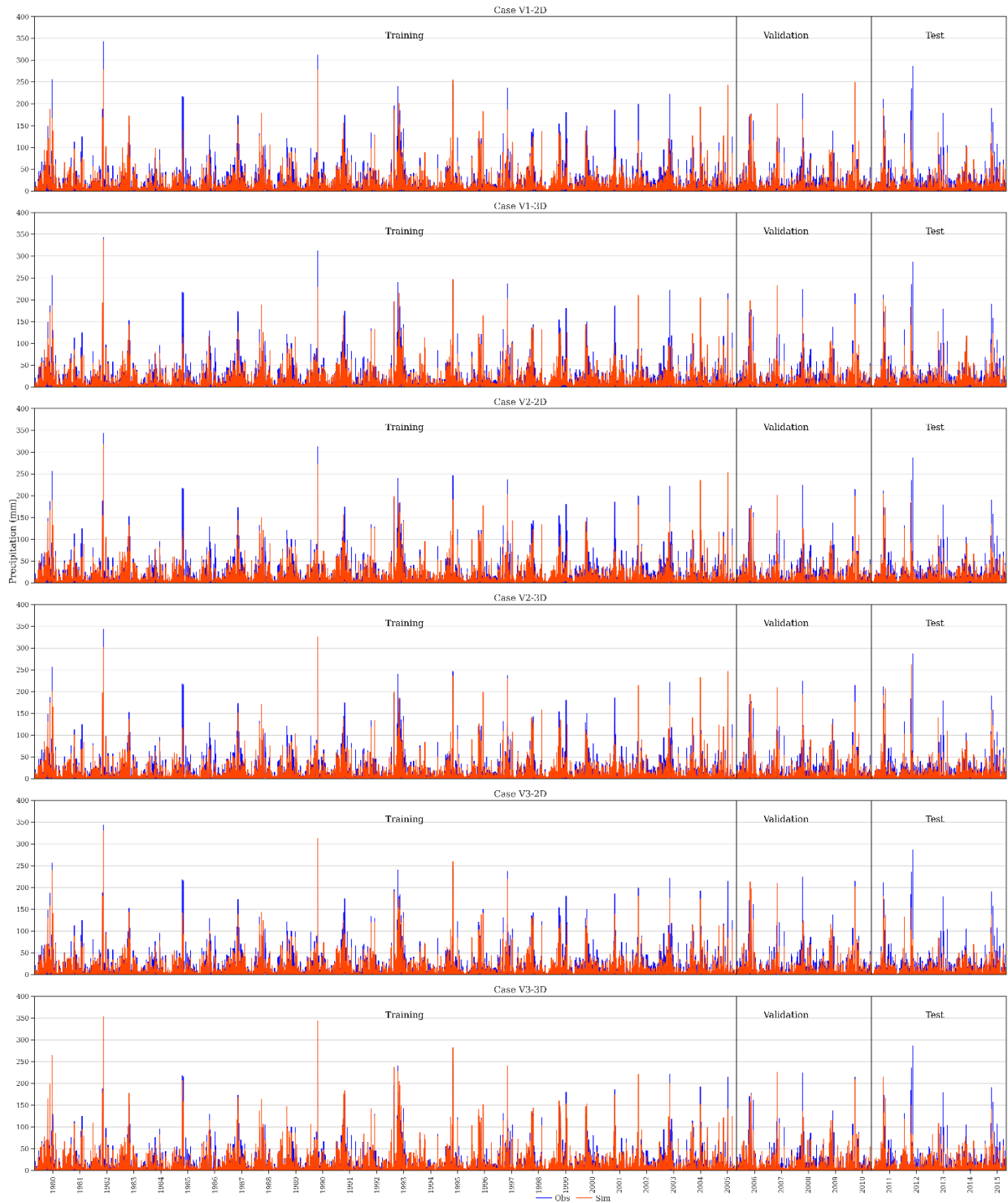

**Figure 7.** Time series plots of estimated (red) and observed (blue) values during the test period when loss was the lowest during the validation period for 2D CNN and 3D-CNN-Vert.

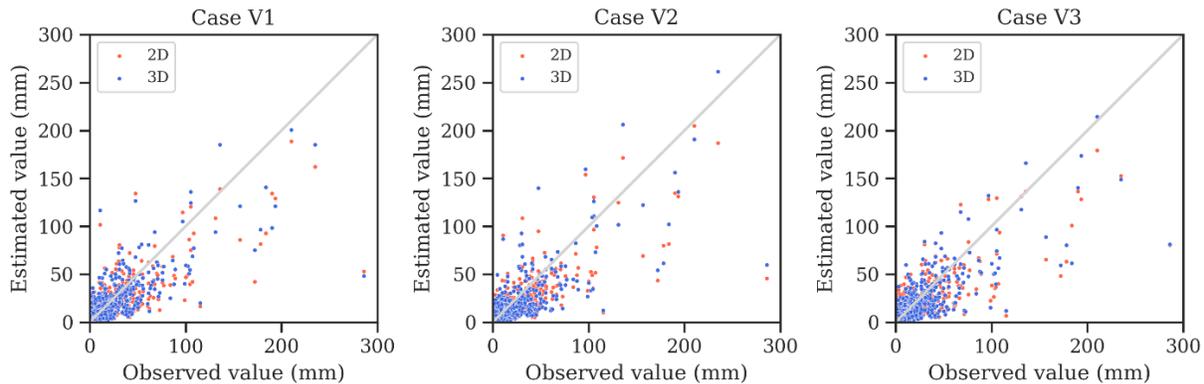

**Figure 8.** Scatter plots of estimated and observed values during the test period when loss was the lowest during the validation period for 2D CNN and 3D-CNN-Vert.

model may also improve the precipitation estimation.

Figures 5-8 show the challenge of accurately obtaining the precipitation peaks. Such difficulty was also found in a previous study by Miao *et al.* (2019). In this study, the use of 3D-CNN-Vert showed some improvements for the peaks in terms of RMSE99 (Table 4). However, the precipitation peaks were still often underestimated. This may be due to an insufficient number of heavy precipitation data. The frequency distribution of precipitation is not flat. This kind of data is called imbalanced data in the model. It is known that it is difficult for a machine learning model to learn from imbalanced data (Johnson and Khoshgoftaar, 2019). The difficulty of replicating the precipitation peaks may be also due to this imbalance. Special treatments for imbalanced data in machine learning are summarized in review papers (e.g., Barandela *et al.*, 2004; Johnson and Khoshgoftaar, 2019). Such treatments may improve estimation accuracy.

## 6. CONCLUSION

This study extended a 2D CNN to a 3D CNN along the time direction (3D-CNN-Time) and along the vertical direction (3D-CNN-Vert) for precipitation estimation from 3D atmospheric time series data. 3D-CNN-Time and 3D-CNN-Vert were compared with the 2D CNN in terms of RMSE, NSE, and RMSE99. A reanalysis dataset, ERA-Interim, was employed to extract atmospheric time series on multiple isobaric surfaces for the inputs. The daily basin-average precipitation at the study watershed was the target.

The results of this study show the potential of both 3D-CNN-Time and 3D-CNN-Vert for improving model accuracy for precipitation estimation compared to a 2D CNN. However, the improvements by 3D-CNN-Time were more limited than those by 3D-CNN-Vert. For instance, 3D-CNN-Vert improved RMSE, NSE, and RMSE99 at most by 0.5 mm, 0.032, and 6.3 mm, respectively for the test period compared to the 2D CNN. 3D-CNN-Vert obtained better RMSE and NSE for the test period although RMSE99 was worsened by 3D-CNN-Vert for the test period in some cases.